%% file: main_vision.tex
\documentclass{article}
\usepackage[utf8]{inputenc}
\usepackage[numbers]{natbib}
\usepackage{graphicx}
\usepackage{hyperref}
\usepackage{amsthm}
\usepackage{authblk}
\usepackage{comment}

\theoremstyle{definition}
\newtheorem{definition}{Definition}[section]

\title{Towards API Testing Across Cloud and Edge}
\author{\mbox{Samuel Ackerman$^1$}, \mbox{Sanjib Choudhury$^2$}, \mbox{Nirmit Desai$^3$}, \mbox{Eitan Farchi$^1$}, \mbox{Dan Gisolfi$^4$}, \mbox{Andrew Hicks$^4$}, \mbox{Saritha Route$^2$}, \mbox{Diptikalyan Saha$^5$}}

\date{%
    $^1$\{samuel.ackerman@, farchi@il.\}ibm.com, IBM Research, Haifa, Israel\\%
    $^2$ \{sanjib.choudhury, saritha.route\}@in.ibm.com, IBM Services, Bangalore, India\\%
    $^3$ nirmit.desai@us.ibm.com, IBM Research, NY, USA\\%
    $^4$ \{daniel.gisolfi@, ahicks@us.\}ibm.com, IBM Systems, NY, USA\\%
    $^5$ diptsaha@in.ibm.com, IBM Research, Bangalore, India\\[2ex]%
    \today
}

\begin{document}
\maketitle

\input{abstract}

\section{Introduction}
API economy is driving the digital transformation of business applications across the hybrid cloud and edge environments. In such applications, micro-services are typically defined and expose a set of API interfaces.  Each micro service serves a specific business need.  By composing the micro services from multiple vendors a greater business value is obtained.   For example, a book store may have an inventory micro-service the buyer can search and a shopping cart micro-service that keeps an inventory of the books the buyer would like to buy.  In addition, a credit validation micro-service is leveraged from another vendor and a book shipping micro-service by yet another vendor.  Together, the micro-services allow the buyer to search the bookstore, decide on the books she would like to buy, and then buy them.  The micro-services may be placed in hybrid Cloud environments across edge locations and multiple Cloud environments and are potentially accessed by the buyer from an edge device.  In such environments, in addition to the functional requirements, communication failures, order of message passing between the micro services, placement of the micro services and failure to react by a micro service (sometimes referred to as circuit breaking) may result in quality issues.  Thus, in the API economy, testing of functional and non-functional requirements should be addressed to create a high quality solution. So, to improve the test coverage, the functional and non-functional requirements should ideally be addressed together in a single test suite.

The challenge of distributed API testing as described above results in the product of several exponential input or triggers spaces that should be covered to adequately test the solution.  For a given API, the space of possible input parameters given API pre-condition space is huge.  In addition the API can run in different processes, and be sequenced in different ways while accessing shared resources.  This results in another huge space of possible test scenarios.  Also, communication may fail or get delayed or some micro-services may go down and fail to respond all together.  This may result in another huge set of possible test triggers.  One approach is to separate these test concerns and test for each concern in  separation. Instead, this work attempts to automatically cover, once the functional tests of the APIs are defined, all other test concerns resulting in a more comprehensive and automated test approach.

To address these challenges, we have created an API test engine called
\emph{Distributed Software Test Kit (DSTK)} discussed in what follows. The main contribution of this vision paper is thus an approach for targeted automated non-functional testing of distributed API-based applications, that focuses around the functional test specification. 
The functional test is designed using Combinatorial Test Design (CTD) \cite{10.1145/2001420.2001451} implemented in IBM's IGNITE Quality Platform (IGNITE)\footnote{https://ibm.biz/BdfGSx}. Once the functional tests are obtained from IBM IGNITE, the engine proceeds to execute the functional tests under various conditions forcing different execution orders, executing different micro-services on different locations across edge and Cloud, injecting network faults, measuring API performance, and introducing changes in API call patterns. Based on the feedback of test results, faults, and drifts in API call patterns,  under-the-hood AI based search techniques are applied to generate new test scenarios for execution to maximize coverage of the non-functional scenarios spanning the functional flows. 

\subsection{Paper Outline}
The rest of this paper is organized as follows. In Section~\ref{sec:dstk}, we describe the novel architecture of DSTK to enable distributed API testing. Next, we formally define a coverage model for non-functional requirements in Section~\ref{sec:coverage}. Equipped with the architecture and the definitions, Section~\ref{sec:generation} describes three unique techniques for test generation: CTD (Combinatorial Test Design), constraint generation, and PDG (Program Dependency Graph). The generated tests then are executed via the orchestration setup described in Section~\ref{sec:orchestration}, including fault-injection and API drift detection.

\section{DSTK Architecture}
\label{sec:dstk}
\begin{figure}[htb!]
\centering
\includegraphics[width=\columnwidth]{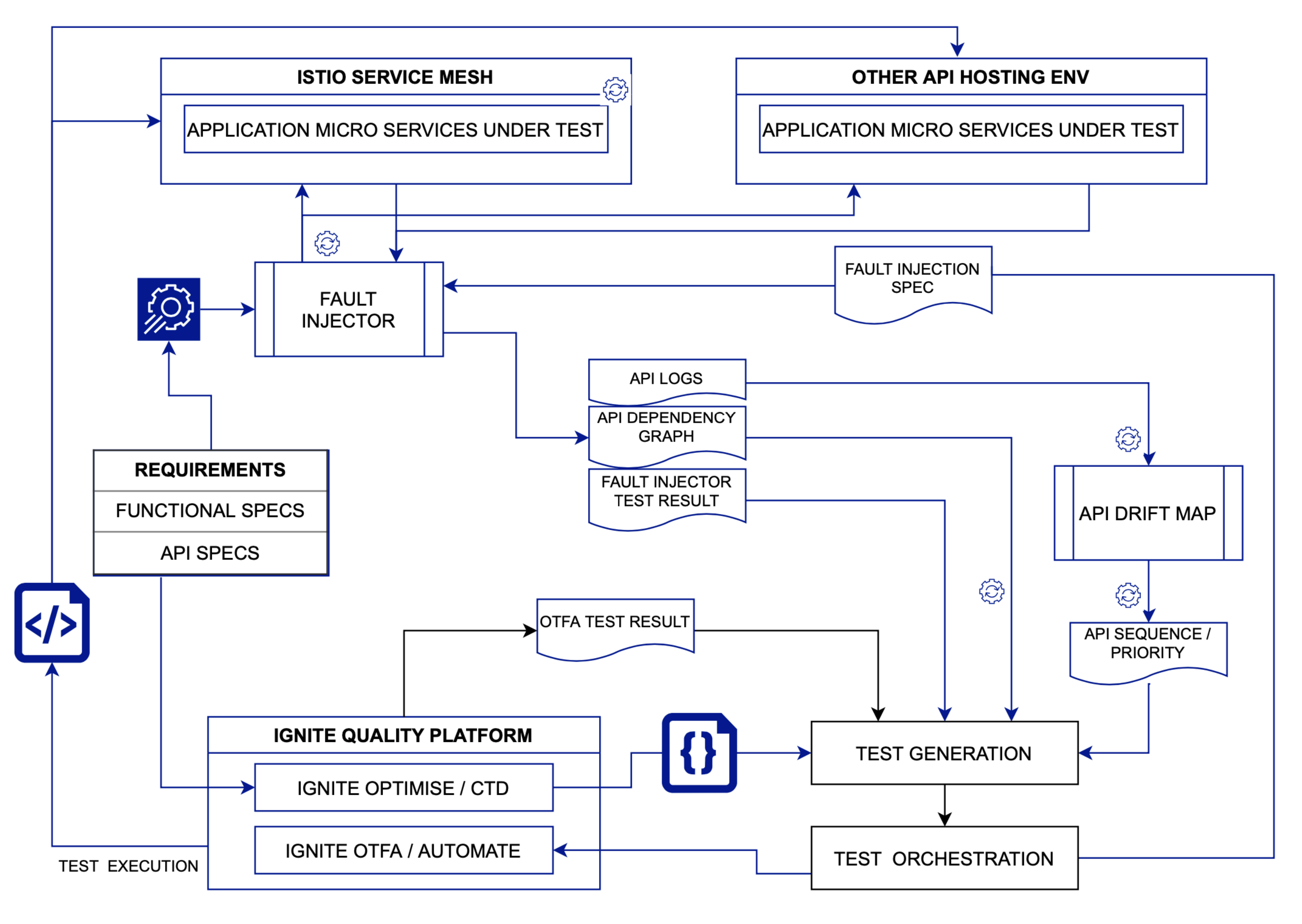}
\caption{Distributed Software Test Kit (DSTK) architecture}
\label{fig:stk_system_context}
\end{figure}

The DSTK itself is implemented as a set of micro-services, each playing a specific role and exposing well-defined APIs. These micro-services can be deployed in a distributed fashion across the hybrid Cloud and edge environments, as needed. Such an architecture is a significant departure from traditional test frameworks and solutions which tend to be monolithic. 
The micro-services based implementation of each of the DSTK components means that a subset of the components can be leveraged, possible with other testing frameworks, in testing a specific application.

Figure~\ref{fig:stk_system_context} shows the key micro-services and how they interact to enable testing of distributed, API-based applications. The application under test (AUT) consists of multiple micro-services, each of which may be deployed in a Cloud or an edge site.  Each site is a Cloud-native environment with a micro-service runtime, e.g., Istio service mesh. 

A tester starts with the functional specs or the API specs of each of the micro-services in the AUT in a standard format, e.g., Open API spec 3.0. Using these specs, the tester leverages the IBM IGNITE Quality Platform (IQP) to automatically generate a set of functional test scenarios via the CTD algorithms implemented within IQP. These functional tests are then automatically translated to executable Cucumber features. 

The DSTK consists of two main component micro-services: (a) a test orchestrator that has the ability to coordinate execution of tests across multiple distributed sites and (b) a test generator that dynamically searches for and generates new reliability tests to improve test coverage while minimizing test duration.

The orchestrator within the DSTK receives the executable Cucumber features from IQP and injects hooks that allow orchestration of the test execution across a distributed environment. The hooks execute one of the following scenarios.

\begin{itemize}
    \item \textbf{Synchronization} Force multiple sites executing a test to wait until all sites reach a certain point. Presence or absence of this hook can alter the execution order of the steps within a test scenario.
    \item \textbf{Fault injection} Introduce a network delay or a complete network breakdown, either within a service mesh or across multiple service mesh instances.
    \item \textbf{Load generation} Alter the frequency of calls made by one micro-service to another to induce performance delays and faults.
    \item \textbf{Placement topology} Alter the topology of what AUT micro-serviced is deployed at what Cloud and edge site.
\end{itemize}

The modified Cucumber features are then executed via the IGNITE Optimized Test Flow Automation (OTFA).  The execution involves making API calls to the AUT micro-services distributed across multiple Cloud and edge sites and collecting API execution logs via the service mesh as well as test results via OTFA.

A drift analysis on the API logs is carried out to observe any changes to the frequency of API calls from a micro-service to another. This in turn can be used to identify changes in workload behaviours.  In addition, the OTFA test results, the API logs, the API dependency graph, the fault injection logs and the drift detection analysis are fed to a search algorithm within the test generator. The main function of the test generator is to calculate test coverage achieved by the test scenarios executed thus far and generate new test scenarios that are likely to improve the coverage significantly while minimizing the test execution time. A variety of search algorithms can be leveraged here, e.g., genetic algorithms, reinforcement learning, simulated annealing, and so on.

Based on the search algorithm, new test cases are generated and handed over to the test orchestrator again to inject necessary hooks and execute via OTFA. This process continues until a desired target coverage is achieved or no improvement in coverage is seen for a pre-defined number of test rounds.

\section{Non-Functional Coverage Model}
\label{sec:coverage}

We next introduce a formal non-functional coverage criterion that focuses on communication breakage.  This is by no means the only possible non-functional coverage criterion of interest.  Other non-functional coverage criterion can be introduced for ordering of the API execution, placement of the micro services.  Information gathered with respect to non-functional coverage models are used to direct the test generation process instead of having a naive chaos test generation.   

We are given a set of API under test $API = \{f_1, \ldots, f_k\}$.   We are also given a set of functional tests, $T = \{t_1, \ldots, t_k\}$.  The execution of $t_i$ results in a series of calls to the API under test.  In addition, a call graph is given $CALL \subseteq API \times API$. There is an edge $(f_i, f_j) \in CALL$ if $f_i$ can call $f_j$ or there is a data dependency between the output of $f_i$ and the input of $f_j$.   

The test orchestrator asks OTFA to call the API under test while simultaneously asking the fault injector to inject specific network faults.  Thus, if as a result of an execution of a functional test $t_l \in T$, $f_i$ calls $f_j$, (thus $(f_i, f_j) \in CALL$), the fault injector assumes control once the call to $f_j$ is made by $f_i$ and before control is passed to $f_j$. As a result, the call can be dropped and delayed, simulating breakage in the communication.  The delayed communication may alter the order in which the internal logic of the API under tests is executed or trigger some timeouts. Thus, we have three events of interest on each edge $(f_i, f_j) \in CALL$ , namely $\{breakage(i, j), delayedHappyPath(i, j), delayedErrorPath(i, j)\}$ which we will define next.

 We assume that $t_l$ can either execute along a happy path realizing the purpose of the transaction, e.g., the delivery of a book to a customer, or it could fail to realize the purpose of the transaction but does so gracefully.  We refer to the latter case as an error path. We also assume that the test $t_l$ has an expected result that can be determined automatically at run time and we can thus determine, automatically, if the error path has occurred or not or if the test failed and we need to debug the potential problem.  Thus, three coverage events are defined as follows
\begin{itemize}
\item breakage - the API call is never realised.
\item delayedHappyPath - random delays are introduced and $t_l$ executes a happy path.   
\item delayedErrorPath - random delays are introduced that cause an error path to be executed by $t_l$.
\end{itemize}

For a given functional test $t_l \in T$ we define, $API_{t_l} \subseteq API$, the set of API that are called by $t_l$.  We consider the sub graph, $CALL_{t_l}$, that is composed by $API_{t_l}$ and edges between them in $CALL$.  For a given functional test $t_l \in T$,  We are thus ready to define our coverage model which we use as a search target in our ML based test generation.   This test generation process is used to reveal problematic handling of error path and timeouts.  

\begin{definition} We say that $t_l \in T$ is circuit breaker covered if the above three events occur for each edge in $CALL_{t_l}$.   We say that $T$ is circuit breaker covered if each $t_l \in T$ is circuit breaker covered. 
\end{definition}

There are two enhancements to the basic circuit breaker coverage model which we detail below.  When running $t_l$ we can change the order of execution of the API calls.  Such perturbation may cause the execution of additional API not executed in the normal non-perturbed run due to race conditions.   We can redefine $API_{t_l}$ to be the set of all API calls that were executed in any run of $t_l$ under the various perturbations introduced by synchronization perturbation.  

Another enhancement of the circuit breaker basic coverage model approaches the issue of $API_{t_l}$ incompleteness discussed in the above paragraph by creating a hierarchy of coverage models.  The formal definition follows.    

\begin{definition} We say that $t_l \in T$ is circuit breaker $i$ covered if each of the three events $\{breakage(i, j), delayedHappyPath(i, j), delayedErrorPath(i, j)\}$ occur for each edge in $CALL^i_{t_l}$.   $CALL^i_{t_l}$ is obtained as follows.  We add interfaces $f_r \in API$ to $API_{t_l}$ for which there is a minimal path from some $f_t \in API_{t_l}$ to $f_r$ that is of length less than $i$ thus getting the set of interfaces $API_{t_l}^i$.  The sub-graph in $CALL$ with vertices in $API_{t_l}^i$ is $CALL^i_{t_l}$. We say that $T$ is circuit breaker $i$ covered if each $t_l \in T$ is circuit breaker i covered. 
\end{definition}

\begin{figure}[htb!]
\centering
\includegraphics[width=\columnwidth]{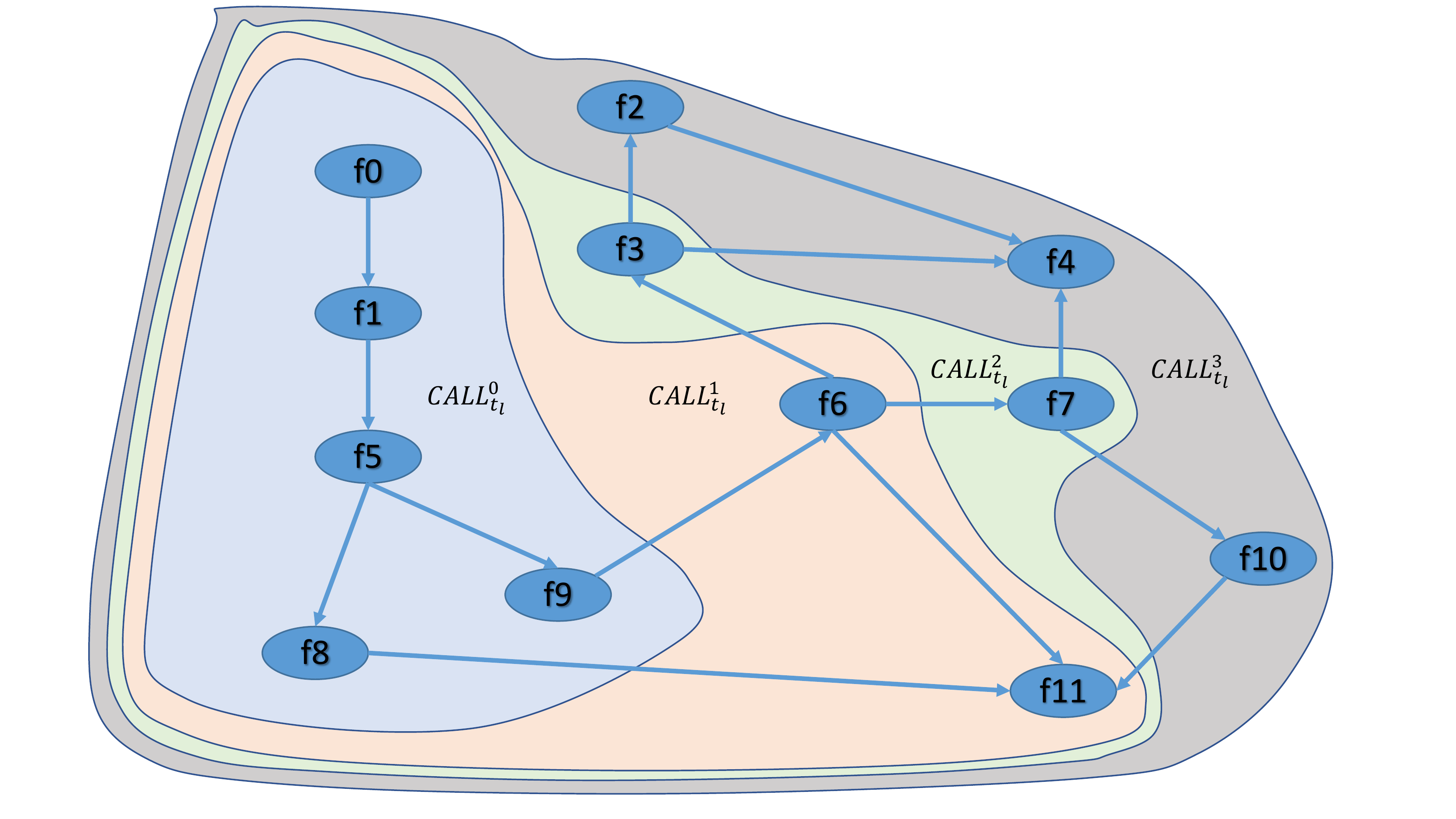}
\caption{Illustration of $API_{tl}^i$ call graph where $API_{t_l} = \{f_0, f_1, f_5, f_8, f_9\}$}
\label{fig:API}
\end{figure}

This coverage hierarchy provides a sequence of harder to cover circuit breakage objective as a function of $t_l$ and $T$ that ends in attempting to cover $CALL$.  As such it helps control the amount of resources spend on testing.  

\section{Approach: Test Generation}
\label{sec:generation}

In this section, we describe exemplar techniques that can be leveraged for automated generation of functional and non-functional test cases, as shown in the "Test Generation" box in Figure~\ref{fig:stk_system_context}. We leverage CTD as the base technique for test generation and refine it by pruning the search space with constraint generation and improving coverage by analyzing the API dependence graph via PDG.

\subsection{Functional Test Generation via CTD}

We use Combinatorial Test Design, CTD\cite{10.1145/2001420.2001451}, to optimally cover input, output and environment states jointly referred to as parameters of API calls. Specifically, we use the IBM IGNITE test automation and optimization platform\footnote{https://www.gartner.com/reviews/market/application-testing-services-worldwide/vendor/ibm/product/ibm-ignite/reviews}.  We follow a light modeling approach as follows.

\begin{enumerate}
    \item Abstract API parameters using sub-domain partition.   For example, if an API input includes a file size we may partition the possible size values to small, medium, and large.  By doing so a Cartesian product of API parameters $P = (P_1, \ldots, P_k)$ is defined. 
    \item Define logical constrain on $P$.  For example, if $P_i$ represents whether a file exists or not and $P_j$ represent permission to create a file and $P_l$ the success of an open operation, we may have a constrain that open fails if the file does not exists or if we do not have the permission to open the file.  This stage partitions $P$ into legal and illegal combinations.
    \item A test $t = (t_1, \ldots, t_k)$ is a legal member of $P$.  $T$ is a set of tests.  $T$ is k-interaction covered if for every $k$ parameters values from $t_{i_1} \in P_{i_1}, \ \ldots, t_{i_k} \in P_{i_k}$ for which there is some legal test in $P$ that includes $t_{i_1}, \ldots, t_{i_k}$, there is a test in $T$ that includes $t_{i_1}, \ldots, t_{i_k}$. In practice 2 to 3 interaction coverage is applied.   The exact choice of what interaction coverage requirements depends on the domain and the test concerns and is thus part of the test design.
\end{enumerate}

By applying the CTD algorithm we obtain a minimal test set $T$ that obtains the desired interaction coverage. As the model include sub-domain partition abstractions (stage one above) we use IGNITE to generate concrete tests that actually call the APIs under test.  Thus a set of functional tests is obtained and is used as a seed to automatically produce by our test endowment, the non functional and edge / distributed compute tests. 

\subsection{Automated constraints generation}

The above modeling process can be partially automated by a variety of approaches described below.






\subsubsection{Automated generation of constraints}


For CTD models that are small and environments that allow automatic test generation, we generate the entire Cartesian product, run every test on the system, and collect the console output for every test. Then with tokenization and clustering, we group and correlate clusters of output and their corresponding test vector. In this manner, no assumptions about the underlying code are made, but rather solely based on code output. This then leaves us with three potential cluster's categories, a cluster for which the output is a bug in the software stack, a cluster for which there is a bug in the test plan, or a cluster which represents an illegal combination that should be captured as a constraint and be removed from the set of test vectors. Once the tester identifies the category of each cluster, we can then derive the constrains automatically from the output, simplifying the process and creating CTD model constrains.  The process also helps define tests expected results, refine and correct the model. Next the CTD optimization algorithm is applied on the refined model (with the added constraints) to obtain an optimized test set.  

\subsubsection{Value of constraint generation for API testing}

Generating restrictions in the manner stated above is extra valuable with testing APIs since the back-end/application layer is hidden from the tester/end-point user. When testing, it becomes exceedingly easy to create the CTD model based on the API specification/user-codes. The CTD model now represents the business logic of the system under test but does not have any depth in how the underlying code is being driven. This is, in essence, 100 present black-box testing, and may miss essential testing hints. Through the use of test console output clustering to identify potential code paths and constraints the use of the API specification for testing is refined and enhanced. 




\subsection{PDG: API Dependence Graphs}

In this section we focus on increasing code coverage of the API tests.  For example, increasing the number program statements or program conditions that are exercised by the tests. API testing platforms usually support the creation of functionally motivated manually defined test cases. As a result, code coverage of those test cases is usually low.  The low code coverage is compounded by a related problem of the time and effort required to  generate feasible test inputs.  Our aim is to automatically generate test cases that execute code paths in the APIs which are not traversed by the manually written test cases.

 The path coverage and feasible test inputs problems stated above can be tackled by using production behaviours of the APIs.  To do so one needs to distinguish the path executed by the manual test cases and production behaviours as well as obtain input data by analysing production inputs. This is easier said than done. Primarily because the APIs are black box implementations and it is not possible to directly instrument the APIs for path profiling. Further, data privacy and confidentiality constraints determine that production data should not be exposed or used in test suites. We describe how we handle these constraints and automatically build out new API tests and test data using a novel technique called PDG (Program or API Dependency Graph).

PDG is a white-box model which captures the behaviour of the black-box APIs. For each API endpoint we build one PDG. Even though PDG can be built to distinguish two sets of test datasets, for the above setup PDG is built on the production data which also address privacy concerns as explained in details below. The PDG based solution has three phases: 1) build, 2) compare, 3) generate. 

Build: The input to the PDG building is the request, response data in the production and optionally the trace data (low cost OpenTracing/Jaeger method can get internal sequence of API calls). If the trace data is present, the algorithm first creates trace-buckets where each bucket represents an acyclic path. If there are $n$ such trace-buckets, $n+1$ models are created. For each such trace bucket, a model is computed - decision tree regression model capturing the relation between request and response corresponding to the trace bucket. Secondly, a decision tree classification model which determines the input condition to go to that particular trace-bucket which becomes a class in the classification model. The build procedure also carefully performs abstraction to the request, response input type space, flattening the JSON input to scalar variables. Also note that there can be an empty trace bucket for the paths which did not encounter any API calls. Note that, the above mentioned PDG only captures the relation between request and responses. We finally add to PDG the constraints related to inputs (statistical properties for each field and association relationship between fields)  which can be inferred from analyzing the production traces. 

Compare:  Once the PDG is built on the production side, it is imported into the test side. Since PDG does not contain explicit values for request response input, such importing is possible. To obtain untested production behaviour, each test input the test suite is mapped to the PDG using the same trace and request input. This let us find which trace bucket is executed and which decision tree path is executed in the model with less error in the output function computation.  A decision tree path is visited if the input does follow the path and matches the output computation. A test case is considered as anomaly with respect to the production trace, if its trace cannot be mapped to the existing trace bucket or it does not follow the decision tree paths in the classification model for this trace bucket or does not match to the output computed to the decision tree path in the regression model within an acceptable threshold.

Generate: This step takes the un-visited paths and gathers the constraints from the decision tree path of the regression model and the path from the corresponding decision tree classification model. The path constraints along with the input constraints are presented to a constraint solver to generate input which will traverse the path. Such generation ensures privacy preserving as concrete inputs from the production is not used to generate such input.

\section{Approach: Test Orchestration}
\label{sec:orchestration}

In this section we focus on the "Test Orchestration" box in Figure~\ref{fig:stk_system_context} and discuss how circuit-breaking and drift detection of a API heat map are used to generate new reliability tests from the seed set of functional tests.  In this way coverage of non functional concerns may be realized using the search algorithms implemented in the test orchestration component.  Note that circuit breaking is used to inject faults and delays and feedback non functional coverage to the orchestration component while the API heat map drift detection is used to analyze the way the APIs are used and indicate changes.  Thus, information flows from and to the orchestration component realizing the test cycles refereed to in the introduction.

\subsection{Circuit breaker}

We discuss the circuit breaker design pattern, its adaptation to testing and introduce a circuit breaker coverage model that is subsequently used to direct our test orchestration and generation.   

\subsubsection{The circuit breaker design pattern}

In Electrical domain, a circuit breaker is an automatically operated electrical switch designed to protect an electrical circuit from damage caused by excess current from an overload or short circuit. Its basic function is to interrupt current flow after a fault is detected and a circuit breaker can be reset (either manually or automatically) to resume normal operation. [source: Wikipedia]
Likewise, software systems has "current" or flow of control and data that is materialized through remote calls to software running in different processes, usually on different machines across a network. These remote calls need to work in conjunction and in sequence to realise and manifest the business functionality.  The functionality is thus distributed across micro services, devices and clouds. 
\begin{figure}[htb!]
\centering
\includegraphics[width=\columnwidth]{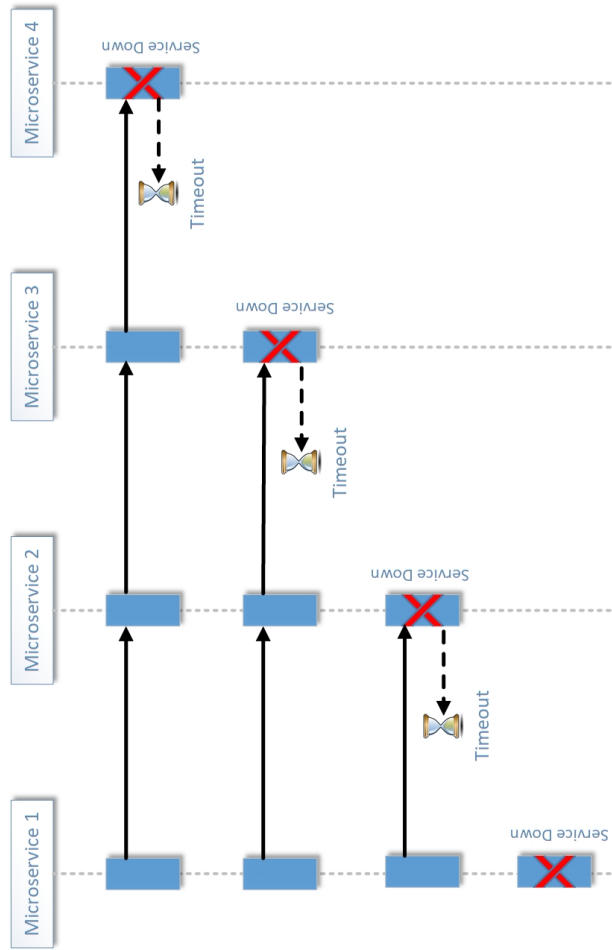}
\caption{\label{fig:SequenceDiagram-NoCircuitBreaker}Sequence diagram of cascading failures.}
\end{figure}

A typical design problem is that remote calls can fail or hang without a response until some timeout limit is reached. If we have many callers on an unresponsive remote call, we can run out of critical resources leading to cascading failures across multiple systems (see figure 3 below).
Michael Nygard in his book Release it popularized the Circuit Breaker pattern to prevent this kind of catastrophic cascade of failures in Software systems. Today, Circuit breaking is an important design pattern for creating resilient micro-service applications. In a nutshell, circuit breaking allows us to write applications that limit the impact of failures, latency spikes, and other undesirable effects of network anomalies.

\begin{figure}[htb!]
\centering
\includegraphics[width=0.7\columnwidth]{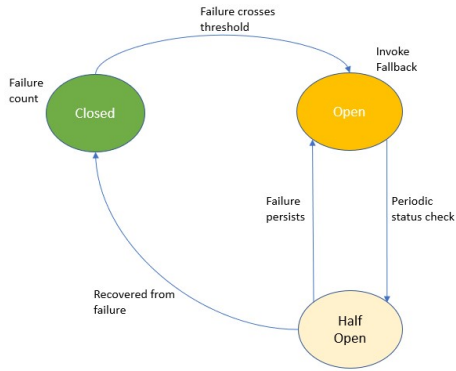}
\caption{\label{fig:CircuitBreaker-Flow}Circuit Breaker flow.}
\end{figure}

The circuit breaker is a proxy that controls flow to an endpoint. If the endpoint fails or is too slow (possibly based on configuration key performance indicators), the proxy will "open" the circuit to the container. In that case, traffic is routed to other containers because of load balancing. The circuit remains open for a pre-configured duration called the sleep window after which the circuit is considered "half-open". The next request attempted will determine if the circuit moves to "closed" (where everything is routed to the endpoint), or it reverts to "open" and the sleep window starts again (see figure 4).

\subsubsection{Utilizing the CB design pattern for testing}

Due to the cascading failure effect and potential systematic effect of end point failure on system behavior it is apparent that end-to-end testing of API and micro services interaction on the cloud and edge is important.  The need to test their behaviour end-to-end arises regardless if the CB design pattern is applied but the CB design pattern infra-structure and specifically its ability to "open" a circuit can be used to achieve the end-to-end testing of the system.  The focus of our end-to-end testing is resilience. 

How do we end-to-end test such a system for resilience? Resilience is the ability of a system to fail gracefully in the face of disruptive events and eventually recover from them. To understand if a system is truly resilient, it is best to measure and understand the resilience of the entire system in the environment where it will run.  To that end we apply fault injection and chaos engineering \footnote{https://principlesofchaos.org/}.

Chaos engineering is the practice of subjecting a system to the real-world failures and dependency disruptions it will face in production. Fault injection is the deliberate introduction of failure into a system in order to validate its robustness and error handling. 
To test a micro-services application, we can inject fault as part of chaos testing. This can be achieved at different layers – for instance, by deleting pods at random, shutting off entire nodes. But failures also happen at the application layer. Infinite loops, broken client libraries - application code can fail in an infinite number of ways.
In this work we have applied Istio fault injection \footnote{https://www.ibm.com/blogs/research/2017/05/amalgam8-istio/}. We have used Istio Virtual Services to do chaos testing at the application layer, by injecting timeouts (delays) or HTTP errors into our services, without actually updating our application code.  We look to bring down services randomly as well as in tandem with a structured analysis of the deployment architecture and the micro services driven by specific test cases that cover functionality and cover edge and distributed compute specific tests and use cases.

\subsection{Performance testing}

Another important feature of test orchestration is performance testing.   To that end, the fault injection component can be utilized to create the appropriate stress and chaos load against the APIs under test.  The test orchestration component and its AI based capabilities can be utilized to then modify the load and enhance it.  We also envision a human-in-the-loop process that helps design the appropriate load on each API. We next describe performance testing in some more details.    

\subsubsection{Our approach}
Let 
$M_1$,\ldots,$M_n$ 
be the number of anomalies contributed by each API $f_1, \ldots, f_n$. Let the response time of each API $f_1,\ldots,f_n$ be 
$R_1$,$R_2$,\ldots,$R_n$ and each throughput be
$T_1$,$T_2$,\ldots,$T_n$. One of the challenges of performance testing is not all seemingly observed anomalies are real anomalies. We would like the testers to refine the loads on the APIs under test based on observed anomalies but we need to filter out spurious anomalies.  We achieve that by training a AI model that   
predicts whether an observed anomaly is actually an anomaly. The model also provides number of real anomalies $M_i$ for each API $f_i$, with response time $R_i$ and throughput $T_i$.   Thus, the overall number of anomalies predicted for the system is $\sum_{m=1}^{n} M_i$ under certain faults injected by some chaos kit. In each iteration we use the above stress results to alert the testers. Based on feedback from testers and developers we change the stress and repeat the above process.

A real life example follows.  A cab facilitator has launched their app which is getting used by several edge devices around the city. The cabs are having IoT devices which are also communicating with the server and among themselves. To have the system conduct the cab booking, invoice generation, tax calculation, and notifying cab IoT device faster, the services have been decoupled and made into multiple micro-services that fit different technologies and frameworks. In production system, if any of these activities fail or lag more than some acceptable average then the performance degradation becomes business critical. 
Given that the fault injector wraps each car micro-service, we are able to specifically control each car fault injection and chaos.  We are thus able to create a non-homogeneous load.  This flexibility is needed for a comprehensive performance test of the system.    


\subsection{API Drift Detection}

\input{heatmap}

\bibliographystyle{plain}
\bibliography{references}
\end{document}

%% file: abstract.tex
\begin{abstract}
API economy is driving the digital transformation of business applications across the hybrid Cloud and edge environments. For such transformations to succeed, end-to-end testing of the application API composition is required.  Testing of API compositions, even in centralized Cloud environments, is challenging as it requires coverage of functional as well as reliability requirements. The combinatorial space of scenarios is huge, e.g., API input parameters, order of API execution, and network faults. Hybrid Cloud and edge environments exacerbate the challenge of API testing due to the need to coordinate test execution across dynamic wide-area networks, possibly across network boundaries.  To handle this challenge, we envision a test framework named \emph{Distributed Software Test Kit (DSTK)}.  The DSTK leverages Combinatorial Test Design (CTD) to cover the functional requirements and then automatically covers the reliability requirements via under-the-hood closed loop between test execution feedback and AI based search algorithms.  In each iteration of the closed loop, the search algorithms generate more reliability test scenarios to be executed next.   Specifically, five kinds of reliability tests are envisioned: out-of-order execution of APIs, network delays and faults, API performance and throughput, changes in API call graph patterns, and changes in application topology.
\end{abstract}

%% file: heatmap.tex
Heat map drift detection is used to help direct the test generation process.  We next discuss our heat map drift detection techniques.  

Let $H$ be an observed sequence of APIs from the set $API$, for instance $H=\{f_1,f_3,f_1,f_2\}$.  The APIs in $H$ represent the order in time in which the APIs call each other.  For instance, the above $H$ means $f_1$ called $f_3$, then $f_3$ called $f_1$, then $f_1$ called $f_2$; in the notation of the directed $CALL$ graph, this mean we have observed instances of the potential edges $H_e=\{(f_1,f_3),\: (f_3,f_1),\:(f_1,f_2)\}$, each of which is in $CALL$, in that order.
Let another observed sequence be $H'=\{f_{10},f_{11},f_{12},f_{10},f_{20}\}$, which likewise means the edge sequence $H_e' =\{(f_{10},f_{11}), \:(f_{11},f_{12}),\:(f_{12},f_{10}),\:(f_{10},f_{20})\}$ is observed.  $H$ and $H'$ can be generated by normal operation or by specific tests $t_i$.

We can assume that $H$ and $H'$ can be modeled as following some random generating procedure, which may be characterized by a probabilistic distribution.  We define drift to mean that a the observed $H'$ is statistically significantly different from some baseline $H$.  We would like to make a sequential determination, that is, we would like to observe, say, elements of $H'$ in sequence (one-by-one or in batches), and be able to determine drift at any moment, without having to wait to observe the entire sequence, which may be long.  A sequential test can reduce risk (saving time, money, etc.) by identifying drift (which may impact business goals) quickly.  

Given a statistical distribution (see below) for the data, the prior distribution is the distribution's parameters under the baseline, while the posterior is the parameter values updated with the observed data.  A likelihood function $L$ is the particular distribution, or model, chosen, which can then be evaluated (have specific parameter values `plugged in').  A commonly-used way of conducting sequential testing is to construct the sequential Bayes Factor (BF), which is the ratio of the likelihood function evaluated at the posterior and prior parameter values.  A BF near 1 means the prior and posterior appear to fit the data similarly, meaning there is minimal or no drift.  When the BF exceeds a given value, we have a corresponding statistical guarantee (pre-determined) of drift from the prior to some other alternative, such as the posterior.

We can think of two alternative ways of modeling an observed sequence $H$. The more naive, but simpler one, is to ingore the fact that the APIs are calling \textit{each other} in the sequence observed, that is, to only consider the set of values $H$ (not the edges).  The statistical assumption here is that when an API $f_i$ is called, this does not change the probability that the next one will be $f_j$, for every possible $(i,j)$ pair.  The observed sequence $H$ can thus be modeled as a realization of a multinomial distribution, which is a categorical distribution with fixed probabilities on each level, but with independent draws each time.  This means that only the total frequency of each $f_i$ is considered, and not the order or frequencies of sequential pairs $(f_i,f_j)$.  This test is illustrated in \cite{LM2020}; the test is still sequential in that the decision is made in time sequence, but that each API call is statistically independent of the ones around it.

A more realistic model is to observe the edge sequences, say $H_e$ vs $H_e'$, and posit a Markovian model, where the edge conditional probabilities $\textrm{Pr}((f_i,f_j))=\textrm{Pr}(f_j\mid f_i)$ are modeled.  While the naive model requires only $k$ (the number of APIs) to be estimated, this typically requires estimating a $k\times k$ matrix $\mathbf{T}$, whose $(i,j)$ entry is the estimate of the edge probability, $\widehat{\textrm{Pr}((f_i,f_j))}$.\footnotemark~  This can be formulated by having the conditionals $\{\textrm{Pr}((f_i,f_j))\},\: j=1,\dots,k$ (i.e., the $i^{\textrm{th}}$ row of $\mathbf{T}$, assuming they are numbered correspondingly), for each $i$ be modeled separately.  The likelihood function is thus the sequence of edge pairs, not the individual $f_i$. 

Since the entries in each row of $\mathbf{T}$ must sum to 1 (this is true of all probability transition matrices), each row can be modeled as a separate Dirichlet distribution with vector of length $k$.  The likelihood can then be evaluated, given a particular posterior (i.e., current estimated) value of $\mathbf{T}$, by multiplying the likelihood of a multinomial of size $n=1$ with the appropriate Dirichlet distribution (row observed), evaluated at the category actually observed.  This can be compared to the likelihood of the sequence under the prior.

We note that we are also interested in the \textit{total frequencies} of pairwise calls as well, not just the probabilities in $\mathbf{T}$.  In the directed  $CALL$ graph, where each node is an API, the magnitude of the edge lengths increases with the number of times each $f_i\rightarrow f_j$ occurs.  This is relevant in the Bayesian paradigm.  Even if two APIs $f_i$ and $f_j$ have the same probabilities of calling the other APIs $f_b,\: b=1,\dots,k$ (i.e., their respective rows in $\mathbf{T}$ are identical), if $f_i$ is called much more frequently than $f_j$, this means we have more \textit{information} about $f_i$'s behavior, and our Bayesian inferences (posterior distribution) will be more certain regarding $\textrm{Pr}((f_i,f_a))$, and the edges $(f_i,f_a)$ will have more magnitude than the respective $(f_j, f_a)$. 

\footnotetext{
Techniques like Bayesian and causal networks can learn which edges need to be estimated and thus avoid having to estimate the full $k\times k$ matrix.
}

%% file: main_vision.bbl
\begin{thebibliography}{1}

\bibitem{LM2020}
{Lindon, Michael} and {Malek, Alan}.
\newblock Sequential testing of multinomial hypotheses with applications to
  detecting implementation errors and missing data in randomized experiments.
\newblock {\em arXiv}, (2), 2020.

\bibitem{10.1145/2001420.2001451}
Itai Segall, Rachel Tzoref-Brill, and Eitan Farchi.
\newblock Using binary decision diagrams for combinatorial test design.
\newblock In {\em Proceedings of the 2011 International Symposium on Software
  Testing and Analysis}, ISSTA '11, page 254–264, New York, NY, USA, 2011.
  Association for Computing Machinery.

\end{thebibliography}
